\documentclass{article}

\usepackage[nonatbib,final]{tackling_climate_workshop_style} % for camera-ready version

\usepackage[utf8]{inputenc} % allow utf-8 input
\usepackage[T1]{fontenc}    % use 8-bit T1 fonts
\usepackage{hyperref}       % hyperlinks
\usepackage{url}            % simple URL typesetting
\usepackage{booktabs}       % professional-quality tables
\usepackage{amsfonts}       % blackboard math symbols
\usepackage{nicefrac}       % compact symbols for 1/2, etc.
\usepackage{microtype}      % microtypography
\usepackage{amsmath}
\usepackage{amssymb}
\usepackage{graphicx}

\newcommand{\Par}{\mathrm{\partial}}

\newcommand{\mbs}[1]{\ensuremath{\boldsymbol{#1}}}

\graphicspath{{img/NIPS2021_fig/}}

\title{Hybrid physics-based and data-driven modeling with calibrated uncertainty
for lithium-ion battery degradation diagnosis and prognosis}

\author{
  Jing Lin, Yu Zhang, Edwin Khoo \\
  Institute for Infocomm Research, A*STAR, Singapore \\
  \texttt{\{lin\_jing, zhang\_yu, edwin\_khoo\}@i2r.a-star.edu.sg} \\
}

\begin{document}

\maketitle

\begin{abstract} % limit of 2000 characters

Advancing lithium-ion batteries (LIBs) in both design and usage is key to promoting electrification in the coming decades to mitigate human-caused climate change.
Inadequate understanding of LIB degradation is an important bottleneck that limits battery durability and safety.
Here, we propose hybrid physics-based and data-driven modeling for online diagnosis and prognosis of battery degradation.
Compared to existing battery modeling efforts, we aim to build a model with physics as its backbone and statistical learning techniques as enhancements. Such a hybrid model has better generalizability and interpretability together with a well-calibrated uncertainty associated with its prediction, rendering it more valuable and relevant to safety-critical applications under realistic usage scenarios.

\end{abstract}

\section{Lithium-ion batteries, electrification and climate change}

By reducing greenhouse gas emissions to mitigate climate change, electrification plays an essential role in distributed energy consumption, such as electric vehicles, and in centralized power grid supply, where energy storage facilities are needed to mediate the mismatch between load requirements and intermittent renewable energy sources such as sunlight, wind, and tide. In such applications, rechargeable lithium-ion batteries (LIBs)~\cite{tarascon_issues_2001} are an increasingly pivotal technology for energy storage and conversion. Therefore, much effort has been devoted in the past decades towards LIB materials development and design improvement~\cite{blomgren_development_2016,kurchin_aced_2020}, modeling~\cite{franco_boosting_2019}, and real-time control~\cite{li_digital_2020}.

A common key aspect underlying these endeavors is
understanding, detecting, and predicting battery degradation~\cite{edge_lithium_2021}, of which the effectiveness will directly impact the performance, durability, safety, and cost of LIBs.
In this project,
we propose the approach of hybrid physics-based and data-driven modeling
for online diagnosis and prognosis,
by which we mean estimation of battery state of health (SOH)
and prediction of remaining useful life (RUL), respectively,
under typical usage patterns.
We would like to emphasize two main threads of model development.
First, we want to minimize reliance on historical usage data in diagnosis and prognosis,
which will enhance the applicability and practicality of the approach.
Second, we make heavy use of known physical knowledge of
both charge/discharge cycling and degradation
to reduce the amount of training data required,
and increase the generalizability as well as interpretability of the modeling approach.

\section{Gaps in past research on battery degradation prediction}

We first categorize current methodologies for SOH estimation, RUL prediction and degradation prognosis along three main dimensions,
all of which affect their relevance to practical usage scenarios, and then review where various lines of past research lie in this landscape.

The first dimension is whether the cycling history of a battery cell is required.
For example, the linear regression model by Severson et al.~\cite{severson_data-driven_2019} requires
the discharge curves of the first 100 cycles for each cell,
which prohibits its generalization and application to a cell at an unknown aging stage.

The second dimension regards whether degradation dynamics is explicitly modeled.
Most purely data-driven SOH predictions take an end-to-end approach,
mapping from past usage~\cite{severson_data-driven_2019,attia_statistical_2021}
or current state directly to RUL.
In contrast,
some researchers attempt to relate incremental capacity fade
to the cell state and usage within a short interval empirically using machine learning (ML)~\cite{sulzer_challenge_2021}.
Such a dynamical model can then be used to project future degradation and RUL
under typical usage patterns.
The latter approach
typically better utilizes the underlying dynamical structure of the system,
yields richer information to be validated,
and more readily adapts to an online estimation setting.
% Moreover, better utilization of the problem structure
% usually relaxes the amount of data needed for model fitting
% and reduces the complexity of the task as well.
% For example, formulating a dynamical system
% is usually much easier than specifying its associated end-to-end flow map.

The third dimension concerns how much prior physical knowledge
is explicitly adopted for SOH prognosis~\cite{sulzer_challenge_2021}.
The prediction models that use minimal physical knowledge are those
purely based on supervised ML and agnostic of any physical principles,
which fit a specified function approximator,
such as a linear regressor or a random forest,
to a set of input-output data by optimization and inference.
For prognosis, the input usually consists of time series of charge, voltage, current and temperature
characterizing the usage history,
while the output is RUL.
Some researchers attempt to map these raw time series directly
to RUL using fully connected or convolutional neural networks~\cite{attia_statistical_2021},
while others leverage physical knowledge about degradation patterns
to engineer some features manually
before feeding them to more conventional algorithms for variable selection and parameter fitting~\cite{severson_data-driven_2019,fermin-cueto_identification_2020}.
Some sequence-based models such as recurrent and long-short-term memory (LSTM) neural networks
are promising for utilizing the raw time-series cycling data available,
but their effectiveness remains to be proven on field data.
Overall, statistical learning theory dictates that
the performance of purely data-driven models will likely deteriorate on unseen data
that are distributed differently from training data,
which is already manifested in the overfitting exhibited by some of the more complex models~\cite{attia_statistical_2021}.
Due to scarcity of laboratory cycling data and noise in field data,
this weakness severely limits such models' applicability to more realistic cell usage.

Lying between purely empirical and physical models
is the family of equivalent circuit models (ECMs).
An ECM uses an electrical circuit with components of resistors, capacitors, inductors, and voltage sources
to mimic the electrical response of a cell to external loads,
without explicitly modeling the underlying reactions and mass transport.
% Therefore, a ECM can model how electric signals vary in battery usage,
% and the resistance and capacity of its components serve as parameters
% that can be fitted to data and characterize SOH.
Due to their simplicity and low computational requirements, ECMs are widely adopted for
real-time SOH estimation in battery management systems~\cite{li_digital_2020,aitio_combining_2020}.
Since all the underlying complicated physics and electrochemistry are lumped into a simple circuit,
an ECM's fidelity can dramatically drop in less normal operating conditions
such as fast charging,
which are exactly the scenarios that cause significant degradation.
% Moreover, an ECM's overly simplified state space
% also prevents it from providing richer diagnosis and prognosis information on degradation
% that is valuable for gaining further insights and adapting usage controlling strategies.

Lastly, purely physical models leverage knowledge on degradation modes
% discovered by intrusive or ex situ measurements,
% such as electron microscopy inspection and impedance spectroscopy,
to formulate dynamical systems that characterize
the degradation evolution~\cite{reniers_review_2019}
in a battery model, such as a single particle~\cite{prada_simplified_2013}
or pseudo-two-dimensional model~\cite{kupper_end--life_2018},
under the framework of porous electrode theory.
Besides their higher computational costs,
these models have not been sufficiently validated by a diverse cycling data set,
and the value they can add to SOH and degradation prediction
has not yet been fully investigated using field data and thus remains unclear to date.
% Moreover,~\cite{reniers_review_2019} also point out the challenge posed by
% the interaction among different degradation modes.

The aforementioned limitations naturally lead to the following question:
putting costs aside,
how far can computational modeling take us towards
reliable online diagnosis and prognosis of battery degradation
that can be adapted to different cell types and usage patterns
without making excessive assumptions about operational histories?
This question is nontrivial because a more complex model
does not necessarily lead to higher fidelity, especially in the regime where multiple highly nonlinear and coupled physical and electrochemical phenomena occur.
We believe it is important to identify the accuracy limit of
uncertainty-calibrated~\cite{mistry_how_2021} computational modeling
in addition to its efficiency.

\section{Proposed hybrid physics-based and data-driven modeling approach}

Here, we propose our initial steps towards studying the accuracy limit
of computational modeling for battery degradation diagnosis and prognosis.

\begin{figure}[!htbp]
\centering
\includegraphics[width=0.7\linewidth]{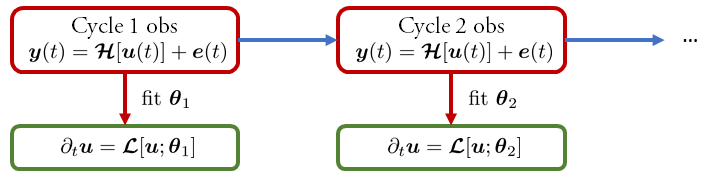}
\caption{
Parameter identifiability and sensitivity study based on a cycling dataset.
Here, $\mbs{y}=\mbs{\mathcal{H}}[\mbs{u}]+\mbs{e}$ is the observation model,
where $\mbs{u}$ is the state vector of the battery model
$\Par_t \mbs{u}=\mbs{\mathcal{L}}[\mbs{u};\mbs{\theta}]$,
$\mbs{\mathcal{H}}[\cdot]$ is the observation operator
that relates the observed quantity $\mbs{y}$ to state $\mbs{u}$,
and $\mbs{e}$ is the observation noise.
Moreover, operator $\mbs{\mathcal{L}}[\cdot]$ characterizes the dynamical battery model
with $\mbs{\theta}$ as the parameters.
}
\label{fig:degrade_param}
\end{figure}

First, we want to investigate the adequacy of various common battery models,
such as variants of ECM, SPM (single particle model), SPMe (single particle model with electrolyte) and DFN (Doyle-Fuller-Newman) models~\cite{Sulzer2021},
for characterizing batteries degraded to different extents.
In particular, we can fit the parameters of these models
to the charge/discharge curves of cells degraded to different stages
from experimental cycling data in~\cite{severson_data-driven_2019}
and obtain the uncertainty associated with parameter estimation
using algorithms for inverse problems, as shown in Figure~\ref{fig:degrade_param}.
The overall model accuracy will indicate whether the parameter space of each model
is sufficiently rich to adequately describe a cell's degradation state,
while the uncertainty will indicate
parameter identifiability~\cite{bizeray_identifiability_2019}
and sensitivity~\cite{park_optimal_2018}.
Moreover, by observing what parameters change across aging and how they vary,
we can diagnose and identify potential underlying degradation mechanisms.

Literature, such as~\cite{sulzer_challenge_2021}, often emphasizes
the ``path dependence'' of battery degradation,
but this is mostly due to an overly simplified representation of the dynamic state of a cell and is not inherent to the underlying physics.
Compared to empirically summarizing the correlation between past usage and future degradation,
we believe pursuing a more refined cell state characterization
that better encodes the impacts of past usage on current SOH
will enable more robust and generalizable prognosis.

\begin{figure}[!htbp]
\centering
\includegraphics[width=0.7\linewidth]{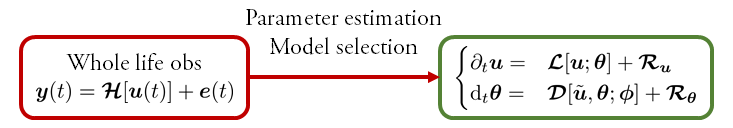}
\caption{
Fitting full battery degradation dynamics to cycling data.
Compared to Figure~\ref{fig:degrade_param},
we also have a dynamical system $\mbs{\theta}$ that models the degradation explicitly
with $\mbs{\phi}$ being its parameters.
$\mbs{\mathcal{R}}_{\mbs{u}}$ and $\mbs{\mathcal{R}}_{\mbs{\theta}}$
denote potential residual data-driven models that can be stochastic processes or ML models.
}
\label{fig:degrade_phys_model}
\end{figure}

Second, as illustrated in Figure~\ref{fig:degrade_phys_model}, we will examine the adequacy of the physics-based dynamic degradation models
reviewed by~\cite{reniers_review_2019},
which describe how certain parameters,
such as porosity, surface area and species diffusivities of the electrodes,
evolve due to side reactions and material fatigue.
We will study whether incorporating these models
will yield similar parameter trends across degradation
to those obtained empirically from the first step.
Again, we will quantify the uncertainty associated with parameter estimation
and check identifiability and sensitivity of the parameters of these degradation models.
We will also attempt model selection jointly with parameter estimation
by, e.g. sparse regression~\cite{rudy_data-driven_2017},
to identify relevant degradation modes from the cycling data automatically.

Moreover, we will also study
potential residual terms
($\mbs{\mathcal{R}}_{\mbs{u}}$ and $\mbs{\mathcal{R}}_{\mbs{\theta}}$
in Figure~\ref{fig:degrade_phys_model})~\cite{aykol_perspectivecombining_2021,willard_integrating_2021}
as certain Gaussian processes or statistical ML models
to account for the patterns of observed degradation data
that cannot be adequately captured by existing physical models.
Unknown functional dependence of parameters on state of charge (SOC) and temperature
can also be modeled this way as model uncertainty.
There have been numerous efforts devoted to scientific ML
for battery degradation~\cite{sulzer_challenge_2021,aykol_perspectivecombining_2021},
and we believe for this case, a physics-based model with ML playing an assistive role
is potentially more accurate and generalizable.
This task may be challenging because of the limited amount of publicly available cycling data,
but even if we cannot identify the model bias accurately,
as long as the magnitude of the model errors is well characterized
by a well calibrated uncertainty,
this can significantly facilitate state and parameter estimation
and yield more reliable predictions.

\begin{figure}[!htbp]
\centering
\includegraphics[width=0.7\linewidth]{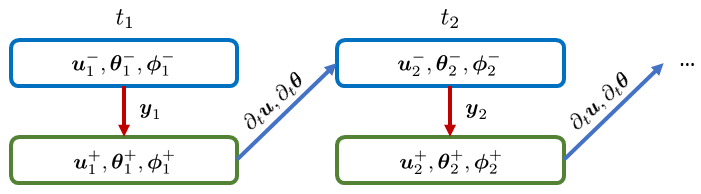}
\caption{Integrating the hybrid modeling approach for battery degradation diagnosis
and prognosis into an online filtering/data assimilation framework~\cite{asch_data_2016}.}
\label{fig:degrade_DA}
\end{figure}

Lastly, in Figure~\ref{fig:degrade_DA}, we will attempt to integrate our hybrid models into
an online filtering/data assimilation framework~\cite{asch_data_2016}
where both state variables and model parameters are jointly updated
by combining model prediction and sequential observation data in an online and principled manner.
Here, we have ample flexibility in choosing what parameters must be inferred online
due to larger variation and sensitivity,
and how frequently they are updated.
While this procedure may be computationally impractical for current battery management systems,
verifying its effectiveness serves as a valuable benchmark
for future model reduction.
The growing power and popularity of cloud computing and internet of things infrastructure~\cite{li_digital_2020}
also makes this approach feasible in the near future.

For our hybrid models to be useful and relevant to the battery community's diverse expertise and requirements, we strive to make them at least easily verifiable by following best practices for reporting sufficient details about the models and associated assumptions and limitations~\cite{mistry_minimal_2021}.

\section{Impacts}

The ability to reliably identify and predict battery degradation is essential
for accelerating battery design iteration and wider adoption of electrification,
which plays a pivotal role in mitigating climate change.
Hybridizing physics-based models and statistical learning
has a high potential for boosting the limit of computational modeling
for online battery degradation diagnosis and prognosis.
Accurate predictions with well-calibrated uncertainty
are valuable for safety-critical applications and
enable further generalization to real field data~\cite{sulzer_challenge_2021}.
With ever-growing cloud storage, computing resources, and wireless connection facilities~\cite{li_digital_2020},
this will also lead to more effective and versatile battery management systems for local control.
Moreover, more effective SOH estimation will likely
unlock more second-life applications such as stationary energy storage~\cite{sulzer_challenge_2021},
making batteries more recyclable and environment-friendly.
Finally, this hybrid modeling endeavor will foster closer collaboration and more idea cross-pollination
between the ML and battery modeling research communities~\cite{baker_fostering_2020}.

\begin{ack}
  This work is supported by Agency for Science, Technology and Research (A*STAR) under the Career Development Fund (C210112037).
\end{ack}

\bibliographystyle{unsrt}
\bibliography{references}

\end{document}